\documentclass[runningheads]{llncs}

\usepackage{eccv}

\usepackage{eccvabbrv}

\usepackage{graphicx}
\usepackage{booktabs}

\usepackage{cuted}
\usepackage{multirow}

\usepackage[accsupp]{axessibility}  % Improves PDF readability for those with disabilities.

\usepackage{hyperref}

\usepackage{orcidlink}

\begin{document}

\title{{\ConDense}: Consistent 2D/3D Pre-training \\ for Dense and Sparse Features \\ from Multi-View Images}

\titlerunning{{\ConDense}}

\author{Xiaoshuai Zhang\inst{1,5\thanks{Work conducted while at Google Research.}},
Zhicheng Wang\inst{5},
Howard Zhou\inst{5},
Soham Ghosh\inst{5},
Danushen Gnanapragasam\inst{5},
Varun Jampani\inst{3,5\footnotemark[1]},
Hao Su\inst{1,4}, 
Leonidas Guibas\inst{2,5}
}

\authorrunning{Zhang et al.}

% \institute{UC San Diego \email{\{zxs,haosu\}@ucsd.edu} \\
% \and
% Stanford University \email{guibas@cs.stanford.edu} \\
% \and
% Stability AI \email{varunjampani@gmail.com} \\
% \and
% Google Research \email{\{nanabyte,howardzhou,sohamg,zhichengw\}@google.com}
% }

\institute{$^1$UC San Diego \, $^2$Stanford University \, $^3$Stability AI \, $^4$Hillbot \, $^5$Google Research}

% project name
% \newcommand{\ConDense}{ConDense}
\newcommand{\ConDense}{\textsc{ConDense}}

% \addtocontents{toc}{\protect\setcounter{tocdepth}{0}}
\maketitle
% \addtocontents{toc}{\protect\setcounter{tocdepth}{0}}

% \definecolor{DarkGreen}{rgb}{0.0, 0.5, 0.0}

\newcommand{\twod}{\texttt{2D}}
\newcommand{\threed}{\texttt{3D}}

\newcommand{\bfo}{\mathbf{o}}
\newcommand{\bfr}{\mathbf{r}}
\newcommand{\bfx}{\mathbf{x}}
\newcommand{\bfd}{\mathbf{d}}
\newcommand{\bfc}{\mathbf{c}}
\newcommand{\bff}{\mathbf{f}}

\newcommand{\bfC}{\mathbf{C}}
\newcommand{\bfF}{\mathbf{F}}
\newcommand{\bfJ}{\mathbf{J}}
\newcommand{\bfI}{\mathbf{I}}
\newcommand{\bfP}{\mathbf{P}}

\newcommand{\calL}{\mathcal{L}}

\newcommand{\boldpara}{\noindent\textbf}

\begin{figure}
  \vspace{-2.5em}
  \centering
  \includegraphics[width=1.0\textwidth]{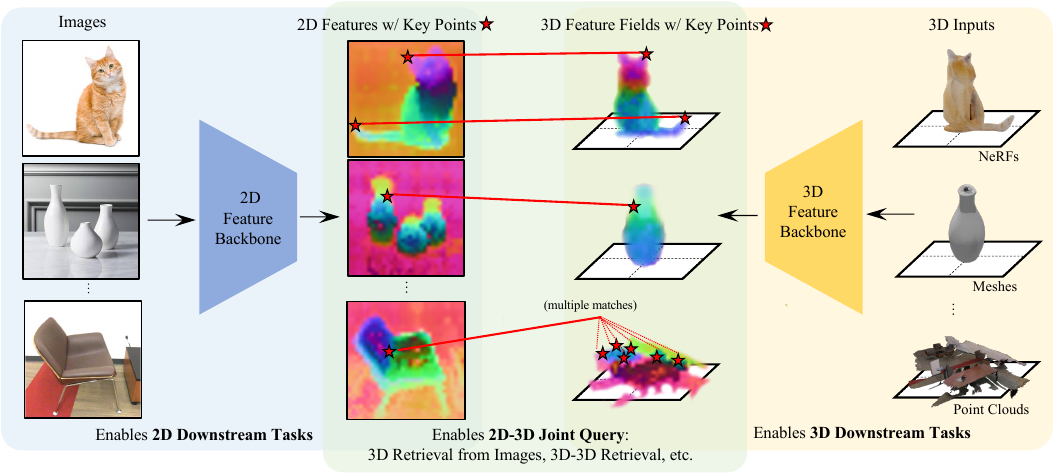}
  \vspace{-2em}
  \captionof{figure}{{\ConDense} extract co-embedded feature for 2D or 3D inputs. The model not only has improved performance over previous pre-training methods but also enables efficient cross-modality, cross-scale queries such as 3D retrieval and duplicate detection.}
  \label{fig:teaser}
  \vspace{-3.5em}
\end{figure}
\begin{abstract}
To advance the state of the art in the creation of 3D foundation models, this paper introduces the {\ConDense}\footnote{\href{https://jetd1.github.io/condense-web/}{Our project page.}} framework for 3D pre-training utilizing existing pre-trained 2D networks and large-scale multi-view datasets. We propose a novel 2D-3D joint training scheme to extract co-embedded 2D and 3D features in an end-to-end pipeline, where 2D-3D feature consistency is enforced through a volume rendering NeRF-like ray marching process. Using dense per pixel features we are able to 1) directly distill the learned priors from 2D models to 3D models and create useful 3D backbones, 2) extract more consistent and less noisy 2D features, 3) formulate a consistent embedding space where 2D, 3D, and other modalities of data (e.g., natural language prompts) can be jointly queried. Furthermore, besides dense features, {\ConDense} can be trained to extract sparse features (e.g., key points), also with 2D-3D consistency -- condensing 3D NeRF representations into compact sets of decorated key points. We demonstrate that our pre-trained model provides good initialization for various 3D tasks including 3D classification and segmentation, outperforming other 3D pre-training methods by a significant margin. It also enables, by exploiting our sparse features, additional useful downstream tasks, such as matching 2D images to 3D scenes, detecting duplicate 3D scenes, and querying a repository of 3D scenes through natural language -- all quite efficiently and without any per-scene fine-tuning.
\end{abstract}
  
\section{Introduction}
\label{sec:intro}
The rapid advancement of 3D computer vision has led to significant breakthroughs in understanding and interpreting the world in three dimensions. However, achieving robust performance across a range of 3D perception tasks is challenging when we try to match the accomplishments of large pre-trained models in the natural language and 2D vision domains. The path to a 3D foundation model is hampered by the relative scarcity of 3D data compared to 2D images, and especially the increased difficulty of obtaining quality annotations in 3D. At the same time, 3D models need to co-exist and communicate with language or language-vision models, if we are to optimally use priors in perceiving, reasoning, and acting on the physical world. Motivated by these considerations we propose a novel approach for large-scale 3D pre-training that leverages the knowledge encoded in extant pre-trained 2D networks, capitalizes on the availability of large-scale multi-view datasets, and learns consistent 2D-3D features.
 
In this paper, we introduce a comprehensive 2D-3D joint training scheme, named {\ConDense}, aimed at extracting co-embedded 2D and 3D features in an end-to-end pipeline. Our approach goes beyond conventional pre-training methods by enforcing 2D-3D feature consistency through 2D-3D consensus. This consistency is established by cross-checking the extracted 2D and 3D features via a ray-marching process inspired by Neural Radiance Fields (NeRFs), ensuring that the learned features align seamlessly in both the 2D and 3D domains.

In addition, {\ConDense} represents the extracted features in two forms: a dense per-pixel representation and a sparse key point-based representation. This dual representation allows us to capitalize on the strengths of both types of features, making \ConDense~versatile and adaptable to a wide range of downstream tasks. Consider, for example, the fact the NeRFs have made the capture of 3D scenes a lightweight and widely available process. We will soon have hundreds of millions of objects and scenes in a NeRF form and the need arises to organize and search large collections of such data -- and in particular to interrogate them using language and image queries. Our sparse key point representation and joint 2D-3D embeddings enable and facilitate such multi-modal cross-domain queries.

Our contributions can be summarized as follows:
\begin{itemize}
    \item We propose {\ConDense}, a novel 3D self-supervised pre-training scheme. By leveraging only large-scale 2D multi-view image datasets and 2D foundation models, we are able to achieve 3D pre-training with state-of-the-art downstream task performance, even when compared to 3D pre-training methods that utilize 3D training data.
    \item Our approach leads to the creation of more consistent and less noisy 2D features, enhancing the quality of existing 2D visual representations, and shows better performance than base models in various downstream 2D tasks.
    \item By learning sparse features jointly with the dense features for both 2D and 3D, we enable several novel tasks such as efficiently matching 2D images to larger-scale 3D scenes, or matching 3D captures of the same scene to each other.
    \item We establish a unified embedding space where 2D, 3D, and other modalities (e.g., natural language prompts) can be jointly queried, enabling efficient matching using either dense or efficient sparse features.
\end{itemize}

To validate the effectiveness of our large-scale pre-training approach, we conduct extensive experiments, showcasing its superior performance in various tasks. Furthermore, our pre-trained model opens up exciting possibilities for downstream applications, such as querying 3D scenes through natural language inputs or efficiently matching 2D images to 3D scenes, all without per-scene fine-tuning.

\section{Related Work}
\label{sec:rel}
\noindent \textbf{2D Representation Learning and Foundation Models.} Initial works on self-supervised 2D representation learning employ various pretext tasks derived from the images themselves~\cite{pathak2016context,zhang2016colorful,carlucci2019domain}, etc. Another line of work adopted discriminative strategies, such as instance classification~\cite{He_2020_CVPR}, treating each image as a unique class and employing data augmentation for training. Recent advances in patch-based architectures, like Vision Transformers (ViTs)~\cite{dosovitskiy2020vit}, revived interest in pretext tasks, particularly inpainting in both image and feature space. Various works find that masked-autoencoders (MAEs) provide strong initialization for downstream tasks~\cite{he2022masked}. However, all these pre-trained features require additional supervised fine-tuning. More recently, foundation models, referring to pre-trained models adept at a broad range of tasks, have seen expansive growth within the vision domain through variants that have successfully adapted to numerous vision-related tasks. Notably, the CLIP model~\cite{radford2021learning} leverages contrastive learning from extensive image-text pairs to achieve zero-shot task transferability. DINO~\cite{DINO,DINOv2} demonstrates the emergence of various desirable properties in its features through self-supervision, facilitating its direct application across diverse visual tasks.

\noindent \textbf{3D Representation Learning and Foundation Models.} Despite advances in 2D representation learning and foundation models, 3D models lag behind greatly due to dataset and architecture constraints. A major line of research proposed various pretext tasks for 3D point clouds~\cite{PointMAE,zhang2021self}. The recent success of ViTs in 2D has also spurred the exploration of their counterparts in 3D domains~\cite{Point-BERT,PointMAE,PointCLIPv2,lin2022meta}. However, all these models require 3D point clouds for pre-training. Most of them are pre-trained on ScanNet~\cite{dai2017scannet} (around 1000 scenes) and ShapeNet~\cite{chang2015shapenet} (around 50k objects, synthetic), and are thus constrained by the limited amount of \emph{real-world} data available.

\noindent \textbf{2D to 3D Feature Distillation and Multi-Modality Embeddings.} With the recent development of large-scale 2D foundation models and multi-modality embeddings (\eg, CLIP for images and languages), many have tried to distill the knowledge learned from these models and extend their application to 3D data formats.
%PointCLIP
PointCLIP and PartSLIP~\cite{zhang2022pointclip,PartSlip} achieves zero-shot point cloud classification and segmentation by projecting point clouds to 2D depth maps and applying the 2D pre-trained models directly. 
%ULIP
OpenShape, ULIP, and ULIP-2~\cite{liu2023openshape,
ULIP,ULIP2} collects text, image, and point cloud triplets and takes advantage of pre-aligned vision-language feature space to achieve alignment among the triplet modalities. Specifically, they fix the visual-language embedding space and only tune their 3D point cloud encoder to achieve this alignment. 
These methods focus on the task of point cloud classification and their design cannot be easily extended to other 3D tasks or 3D input formats.
%OpenScene
Several methods have been proposed for dense feature encoding in 3D~\cite{OpenScene,PointCLIPv2,Uni3D}. For example, OpenScene~\cite{OpenScene} trains on point cloud-grounded multi-view datasets and learns a 3D point cloud network from multi-view aggregated 2D features. Like PointCLIP and ULIP, OpenScene requires 3D point clouds for training, which are scarce and hard to collect in the real world on a large scale, when compared to multi-view images. These works also re-use the embedding space from the 2D foundation model and only distill the 3D encoder.

%DFF, LeRF
More recently, Neural Radiance Fields (NeRFs)~\cite{mildenhall2021nerf} and numerous subsequent follow-ups~\cite{barron2022mipnerf360,zhang2022nerfusion} have gained great success in novel view synthesis. NeRF has the property of aggregating information across views. Several recent works leverage this property to improve the quality of semantic segmentation~\cite{LeRF,nerflets,semanticnerf}. Many works distill features (\eg, DINO~\cite{DINO} and CLIP~\cite{clip}) into 3D and demonstrate they can be used for downstream tasks such as natural language-based query. These works require per-scene distillation and optimization.
%FeatureNeRF
FeatureNeRF~\cite{FeatureNeRF} proposed to distill features from 2D foundation models to 3D space via generalizable NeRFs~\cite{PixelNeRF,chen2021mvsnerf}. Through distillation, the learned model can lift any 2D images to continuous 3D semantic feature volumes. However, the pipeline serves more as a 2D-to-3D lifting technique and cannot handle native 3D data directly.

\noindent \textbf{NeRF for Perception.}
The integration of Neural Radiance Fields (NeRF) into various discriminative perception tasks such as classification, detection, and segmentation has also been explored~\cite{nerflets,nesf}. These methods typically follow a reconstruction-then-detection pipeline by creating NeRFs from multi-view image data first and then designing task-specific networks and loss terms to tackle each perception task, and many of them work in a scene-by-scene manner and require re-training and optimization for each new scene. More recently, NeRF-Det~\cite{nerfdet} incorporates generalizable, feature-conditioned NeRF, and 3D detection pipelines to achieve efficient detection performance with no per-scene fine-tuning. Our work is inspired by these ideas while developing further by joining forces with 2D foundation models and creating a pre-training pipeline that is native to both 2D images and 3D formats.

\noindent \textbf{Querying 3D Data.}
A variety of works have explored 3D to 3D similarity queries in pre-deep learning period, mostly at the object level, by constructing human-designed whole object features encoded as Euclidean embeddings or as distributions~\cite{osada2001matching,chen3d}. Later some works explored learned embedding spaces, as well as co-embeddings of images and 3D models~\cite{yangyan2d3d}. Inspired by the bag-of-words paradigm in image search, ``bags-of-features'' have also been investigated in 3D to 3D search, for example~\cite{bronstein2011shape}. However, none of these approaches is integrated into a multi-task framework, as we aim to do here and they are largely focused on object retrieval, not scenes.

\section{Preliminaries}
\noindent\textbf{Neural Radiance Fields (NeRFs)}~\cite{mildenhall2021nerf} offer a novel representation of 3D scenes, capturing continuous volumetric scenes as neural networks. We briefly describe the mechanism of the NeRF and refer to~\cite{mildenhall2021nerf,barron2022mipnerf360} for details about related NeRF models. A NeRF $\mathcal{F}$ maps a 3D coordinate $\bfx = (x, y, z)$ and a viewing direction $\bfd = (\theta, \phi)$ to a color $\bfc = (R, G, B)$ and density $\sigma$ -- $\mathcal{F} : (\bfx, \bfd) \mapsto (\bfc, \sigma)$,
where color $\bfc$ is related to both point location $\bfx$ and viewing direction $\bfd$, recording the local appearance information, and density $\sigma$ is only related to point location $\bfx$, recording the local geometry information.

The rendering of a 3D scene from a 2D perspective is formulated as a volume rendering problem. Given a ray $\bfr(t) = \bfo + t\bfd$, where $\bfo$ is the camera origin and $t$ is the distance along the viewing direction $\bfd$, the color $\bfC(\bfr)$ of the ray is computed as:
\begin{align}
\bfC(\bfr) = \int_{t_n}^{t_f} T(t) \sigma(\bfr(t)) \bfc(\bfr(t), \bfd) \, dt, \:\:\:\:
T(t) = \exp\left(-\int_{t_n}^t \sigma(\bfr(s)) \, ds\right),
\end{align}
where $T(t)$ is the accumulated transmittance along the ray from $t_n$ (near bound) to $t_f$ (far bound). In practice, discretized approximations are used to evaluate this integral. The process is also called ray marching, which can be used as a general tool to render from any 3D feature field $\bff$ and get a 2D-projected feature map as shown in many previous works~\cite{semanticnerf,nesf,nerflets}. The property can be used to bridge the 3D feature field of a scene with its 2D feature maps and is the basis of our 2D-3D consensus pipeline.

\noindent\textbf{2D Foundation Models} are deep learning models that have been extensively pre-trained on large-scale image datasets. Let $\mathcal{G}$ be a 2D foundation model that maps an input image $\mathbf{I}$ to a feature representation $\mathbf{F}$: $\mathcal{G} : \mathbf{I} \mapsto \mathbf{F}$. The feature representation $\mathbf{F}$ is a high-dimensional vector that captures the essential properties such as geometry and semantic information of the input image $\mathbf{I}$. Of particular interest, our work leverages DINO~\cite{DINO, DINOv2}, a self-supervised learning approach for visual representation trained on large-scale image datasets. Its latest version, DINOv2, excels in capturing both fine-grained details and global contextual information from images without any labeled data. By initializing the 2D encoder from DINOv2, we tap into a robust source of information-rich 2D dense features, and can thus kick-start our 3D encoder from 2D-3D knowledge distillation.

\section{Method}
\begin{figure*}[t]
  \centering
  \includegraphics[width=0.96\textwidth]{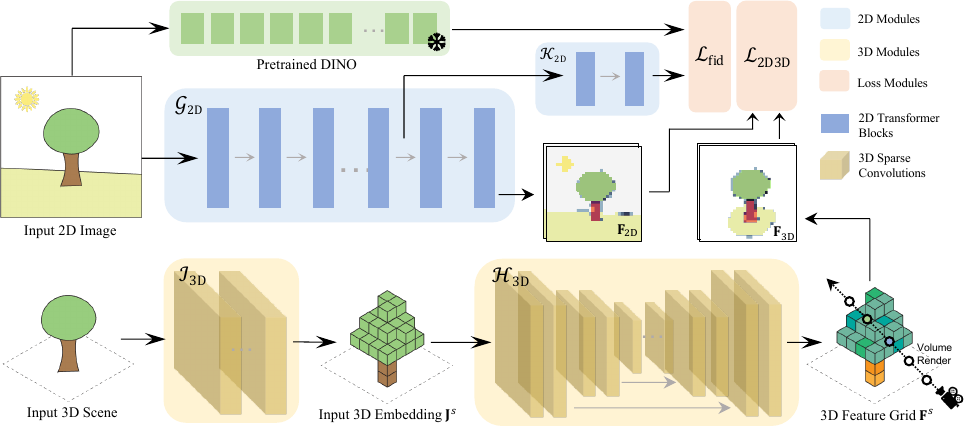}
   \caption{Dense feature encoding: the 3D encoding module $\mathcal{G}_{\threed}$ is composed of a swappable input processing head $\mathcal{J}_\threed$ and a common 3D reasoning backbone $\mathcal{H}_{\threed}$. $\mathcal{J}_\threed$ maps input 3D scenes of various formats into a feature $\bfJ^s$ in a unified 3D embedding space. $\mathcal{H}_{\threed}$ turns $\bfJ^s$ into a 3D feature grid $\bfF^s$. Through interpolation on $\bfF^s$ and volume rendering, a 3D-projected feature map $\bfF_{\threed}$ can be obtained and compared with a 2D dense feature map $\bfF_{\twod}$, extracted from the 2D encoding module $\mathcal{G}_{\twod}$. The resulting 2D-3D consensus loss $\calL_\texttt{2D3D}$ is used as a self-supervision signal. An additional 2D fidelity loss $\calL_\texttt{fid}$ is introduced to make sure that the 2D-3D consensus optimized 2D feature $\bfF_{\twod}$ does not deviate too much from the original 2D feature in order to retain some of its semantics and visual richness.}
   \label{fig:method}
\end{figure*}
An overview of our approach is illustrated in Fig.~\ref{fig:method} (dense feature encoding) and Fig.~\ref{fig:method_keypoint} (key point prediction). Our model is composed of two branches, encoding 2D and 3D information, respectively (Sec.~\ref{sec:method_2d} and Sec.~\ref{sec:method_3d}). Both branches encode features in two forms -- dense format and key-point-based sparse format (Sec.~\ref{sec:method_kp}). During training, we use paired 2D-3D inputs in the form of multi-view images and the corresponding NeRF scene. The information extracted in 2D and 3D branches is compared via 2D-3D consensus (Sec.~\ref{sec:method_2d3d}) so that information can flow both ways.

\subsection{2D Encoding}
\label{sec:method_2d}
The 2D encoding branch ($\mathcal{G}_{\twod}$) of the {\ConDense} framework is crucial for extracting rich visual features from multi-view images, which later are learned synergistically with the 3D encoding module. We follow the network architecture of DINOv2~\cite{DINOv2} to use ViT~\cite{dosovitskiy2020vit} as the base for our 2D encoder and load the pre-trained DINO weights for the initialization of our 2D branch. The ViT architecture takes as input a grid of non-overlapping contiguous image patches of resolution $N \times N$. In this paper, we use $N = 14$ (``/14'' ViT models). The patches are then passed through a linear layer to form a set of embeddings. Following previous works~\cite{dosovitskiy2020vit,DINO}, an extra learnable \texttt{[CLS]} token is added to the sequence to aggregate global information. These patch tokens and the \texttt{[CLS]} token are fed to the standard transformer blocks and updated by the attention mechanism.

For any given input image $\mathbf{I}$, the 2D branch generates a dense feature map $\mathbf{F}_\twod = \mathcal{G}_\twod(\mathbf{I})$. This feature representation $\mathbf{F}_\twod$ is a high-dimensional vector encoding various essential attributions of the input image, and can already be used out-of-the-box due to its pre-training on large-scale image datasets. We will later show that combined with our 2D-3D joint training, the feature branch can be further improved for various downstream tasks.

\subsection{3D Encoding}
\label{sec:method_3d}
The 3D encoding branch ($\mathcal{G}_\threed$) of the {\ConDense} framework is aimed at extracting a 3D feature field from various data formats. It is designed to be composed of two parts: $\mathcal{G}_\threed=\mathcal{H}_{\threed}\circ\mathcal{J}_{\threed}$, where $\mathcal{J}_{\threed}$ is the input processing head and $\mathcal{H}_{\threed}$ is the actual 3D reasoning backbone. Different 3D data formats have their individual input processing heads, but they share a common 3D reasoning backbone $\mathcal{H}_{\threed}$. The major 3D inputs we are dealing with are NeRF models, while we also support other data formats such as point clouds. Here we detail the full pipeline for 3D feature field reasoning from NeRF data.

\boldpara{Grid Sampling from NeRF.} Given any learned NeRF function $\mathcal{F}: (\sigma, \bfc)$, we sample uniformly on a 3D lattice within the normalized scene bounding box $[-1,+1]$, with spacing $\epsilon$ between samples:
\begin{align}
    \Sigma^\text{s} &= \{\sigma(x), \text{s.t.}\ \bfx \in [-1:\epsilon:1]^3\}, \\
    \bfC^\text{s} &= \{\bfc(x,\bfd), \text{s.t.}\ \bfx \in [-1:\epsilon:1]^3, \bfd \in \mathcal{D}\},
\end{align}

\noindent where $\mathcal{D}$ is a predefined set of directions aiming to capture as much local appearance information as possible. In order to reduce computation costs, we use the density field from the input NeRF to sparsify these grids. Specifically, we calculate sample opacity by $\alpha(\bfx)=1-\exp(-\sigma(\bfx))$ due to the canceled-out spacing term $\delta_t$ in the regular grid~\cite{nerfdet}, and filter out the point samples $\bfx$ with $\alpha(\bfx)<\theta$. After sparsification, the evaluated sigma value and color values are concatenated for each grid sample and fed into the input processing head, which is a small 3D sparse convolutional network:
\begin{align}
    \bfJ^s = \mathcal{J}_{\threed}^\text{NeRF}(\texttt{Concat}(\Sigma^\text{s}, \bfC^\text{s})).
\end{align}

Here $\bfJ^s$ serves as the input embedding for the 3D reasoning backbone, while different input processing heads take care of mapping data from different input sources to this input embedding space. For point cloud inputs, we first voxelize the data before feeding it into a small 3D network. Please check the appendix for more details.

\boldpara{3D Spatial Reasoning.} We apply a 3D UNet~\cite{cciccek20163d} implemented with sparse convolution blocks to obtain a 3D feature grid $\bfF^s$ from the aforementioned input embedding:
\begin{align}
    \bfF^\text{s} &= \mathcal{H}_{\threed}(\bfJ^s).
\end{align}

The module enables reasoning in 3D space. To obtain the feature for an arbitrary 3D query point $\bfx \in \mathbb{R}^3$, we interpolate within the feature grid $\bfF^s$ with a trilinear interpolation operator:
\begin{align}
    \bff_\threed(\bfx) &= \mathtt{TriLerp}(\bfx, \bfF^s).
\label{eq:f_3d}
\end{align}

\subsection{2D-3D Consensus with 2D Fidelity}
\label{sec:method_2d3d}
With the proposed 2D and 3D encoders, our model can generate co-embedded dense features given 2D or 3D inputs. During training, we use paired data of multi-view 2D images $\{\bfI_k\}$ and their corresponding learned NeRF $\mathcal{F}: (\sigma, \bfc)$ to jointly train the 2D and 3D branches.

Specifically, for each scene, we first generate the 3D feature field $\bff_\threed$ as detailed in Sec.~\ref{sec:method_3d}. Based on this feature field and the scene density $\sigma$, we can adapt the rendering equation to render 3D-projected feature maps as in ~\cite{semanticnerf,nesf}:
\begin{align}
\bfF_\threed(\bfr) = \int_{t_n}^{t_f} T(t) \sigma(\bfr(t)) \bff_\threed(\bfr(t)) \, dt.
\end{align}

In the meantime, we generate the 2D feature map with our 2D branch: $\bfF_\twod = \mathcal{G}_\twod(\mathbf{I}_k)$ and adopt the consistency loss between the 3D rendered feature map $\bfF_\threed$ the 2D-originated counterpart $\bfF_\twod$ with $L_2$ loss:
\begin{align}
    \calL_\texttt{2D3D}(\bfF_\twod, \bfF_\threed) = \sum_{\bfr\in\mathcal{R}}||\bfF_\twod(\bfr)-\bfF_\threed(\bfr)||^2_2.
\end{align}

$\mathcal{R}$ denotes all camera rays in the multi-view image set of the scene that hit at least one active voxel in the 3D feature grid $\bfF^s$. The loss encourages information to flow both ways -- the 3D branch could learn to generate useful 3D feature fields from the 2D multi-view supervision, and the 2D branch could also benefit from consistent underlying 3D geometry and learn to extract less noisy, multi-view consistent, and 3D-informed features.

Due to the biased and scarcer nature of the existing multi-view image datasets, if we optimize the networks based only on this 2D-3D consistency loss, the feature quality may degrade due to trivial solutions and biased data distribution. To prevent this, we propose to insert an additional output head called 2D fidelity head $\mathcal{K}_\twod$ before the second-to-last transfer block (see Fig.~\ref{fig:method}), and apply the 2D fidelity loss to keep its output $\bfF_\twod^\texttt{fid}$ from deviating too much from the original DINOv2~\cite{DINO} feature:
\begin{align}
    \calL_\texttt{fid}(\bfF_\twod^\texttt{fid}) = ||\bfF_\twod^\texttt{fid} - \bfF_\twod^\texttt{t}||^2_2,
\end{align}
\noindent where $\bfF_\twod^\texttt{t}$ is the pre-trained DINOv2 feature. No ground-truth labels are used in this loss, and this term can be applied on any natural image collection. We adapt ImageNet-21k~\cite{ridnik2021imagenet} in our pipeline.

\subsection{Key Point Extraction}
\label{sec:method_kp}
\begin{figure}[t]
  \centering
  \includegraphics[width=0.6\textwidth]{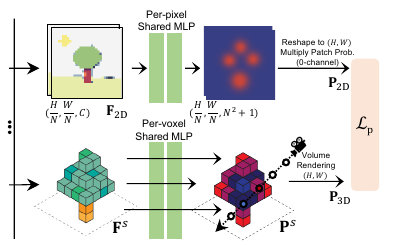}
   \caption{Key point prediction: key points are detected in both 2D ($\bfP_\twod$) and 3D ($\bfP_\threed$) based on the existing feature backbones. The 2D-3D key point loss $\calL_\texttt{p}$ is used as a self-supervision signal.}
   \label{fig:method_keypoint}
\end{figure}
With the proposed 2D and 3D encoders and the joint training scheme, our model can generate co-embedded dense features given any input 2D image or 3D scene. This property is desired since it enables possible applications to query among 2D, 3D, and other modalities. To further facilitate these applications, we add support for sparse key point detection in both 2D and 3D based on the existing feature backbones for efficient queries across scene scales.

As shown in Fig.~\ref{fig:method_keypoint}. To detect key points in 2D images, we follow a similar formulation as in~\cite{detone2018superpoint} and decode the 2D backbone feature $\bfF_\twod$ into the full image-resolution interest point possibility map $\bfP_\twod$ with 2 MLP layers with a softmax output head (noted as $\mathcal{M}_\twod$). Please refer to~\cite{detone2018superpoint} and our Appendix for more in-depth details. For the 3D branch, we similarly use 2 MLP layers with a ReLU output head (noted as $\mathcal{M}_\threed$) to decode the key point possibility on the 3D feature grid $\bfF^s$, and the possibility maps are rendered and compared between 2D and 3D using the aforementioned 2D-3D consensus scheme:
\begin{align}
    \bfP_\twod &= \mathcal{M}_\twod(\bfF_\twod), \\
    \bfP_\threed(\bfr) &= \int_{t_n}^{t_f} T(t) \sigma(\bfr(t)) p_\threed(\bfr(t)) \, dt, \\
    \bfP^s &= \mathcal{M}_\threed(\bfF^s),\ p_\threed(\bfx) = \mathtt{TriLerp}(\bfx, \bfP^s), \\
    \calL_\texttt{p}&(\bfP_\twod, \bfP_\threed) = \sum_{\bfr\in\mathcal{R}}||\bfP_\twod(\bfr)-\bfP_\threed(\bfr)||^2_2.
\end{align}

During test time, 3D key points are selected directly from opaque 3D grid samples. We multiply the opacity value by the predicted key point possibility and use $\alpha(x)\times p_\threed$ as the selecting criteria for the 3D key points.

We argue that the joint 2D-3D key point detection not only helps enable various query tasks based on dense feature backbones (Sec.~\ref{sec:exp_xmod}) but also serves as a useful technique to improve the overall feature quality (Sec.~\ref{sec:abl}). 

\subsection{Training Details}

\boldpara{Loss Terms.} The final loss $\calL$ is given by:
\begin{align}
    \calL = \lambda_\texttt{2D3D}\calL_\texttt{2D3D} + \lambda_\texttt{fid}\calL_\texttt{fid} + \lambda_\texttt{p}\calL_\texttt{p},
\label{eq:loss_final}
\end{align}

\noindent where $\lambda_\texttt{2D3D}$, $\lambda_\texttt{fid}$, and $\lambda_\texttt{p}$ are scalars adjusted throughout the training process. Check the appendix for more details.

\boldpara{Datasets and Model Details.} For all experiments included in the main experiments, we use MVImgNet~\cite{yu2023mvimgnet}, ScanNet~\cite{dai2017scannet}, and RealEstate10k~\cite{zhou2018realestate10k} as our multi-view pre-training datasets. MVImgNet is an object-centric dataset containing a total of 6.5 million frames from more than 200k video captures of diverse objects. ScanNet and RealEstate10k are indoor scene-scale datasets each containing a diverse set of scene captures in the form of video clips. Though other modalities are provided in some of these datasets (\eg point clouds, semantic labels, etc.), we only use the posed images in pre-training. The scenes are fit into MipNeRF-360~\cite{barron2022mipnerf360} models individually before being used for pre-training. ImageNet-21k~\cite{ridnik2021imagenet} is used to provide image samples for the 2D fidelity loss in addition to the multi-view datasets. ImageNet-21k is the superset of the commonly used ImageNet-1k dataset and contains more than 14 million images. We utilize Vision Transformers (ViT-g/14) as the backbone for the 2D branch and use 8 A100 GPUs for training. See the Appendix for more details on data pre-processing and model hyper-parameters.

\boldpara{Training Scheme.} We bootstrap the full training process of {\ConDense} in four stages. First, the 2D feature backbone $\mathcal{G}_\twod$ is initialized from DINOv2~\cite{DINO} pre-trained weights. We then freeze its weights and fit the 2D key point detector MLPs ($\mathcal{M}_\twod$) by enforcing the interest point heatmap predictions to a pre-trained, frozen SuperPoint~\cite{detone2018superpoint} model. Then both $\mathcal{G}_\twod$ and $\mathcal{M}_\twod$ are kept frozen, while the 3D branch modules $\mathcal{G}_\threed$ and $\mathcal{M}_\threed$ are optimized from $\mathcal{L}_\texttt{2D3D}$ and $\mathcal{L}_\texttt{p}$. In this stage, we distill the knowledge from the 2D foundation models to kick start the learning of 3D modules. For the final phase, we unfreeze all modules and jointly train all 2D and 3D modules with the loss terms defined in Eq.~\ref{eq:loss_final}. 

\section{Experiments}
In this section, we present extensive evaluations of our models on 1) 3D tasks including 3D classification, and 3D segmentation; 2) 2D image understanding tasks; and 3) Cross-Modality scene queries. In all experiments, we freeze weights for the feature backbones $\mathcal{G}_\twod$ and $\mathcal{H}_\threed$ unless otherwise stated. Due to the page limit, we only include the most common benchmarks in the main paper, please check the appendix for more experiments including 3D detection, 2D retrieval, and 2D depth estimation.

\subsection{3D Tasks}
For point-cloud-based 3D tasks, we use the 3D feature backbone $\mathcal{H}_\threed$ \emph{out-of-the-box} and freeze its weights, while training a point cloud input head $\mathcal{J}_{\threed}^\text{PC}$ with 4 sparse convolutional layers. For a 3D point $\bfx$, we fetch the interpolated features in the 3D feature grid $\bfF^s$ as the point feature (Eq.~\ref{eq:f_3d}). Depending on the actual 3D tasks, different output heads could be added to further process these point features.

\noindent\textbf{3D Classification} We follow the testing protocols in the previous works~\cite{ULIP,ULIP2} to evaluate on ModelNet40~\cite{wu2015modelnet} and ScanObjectNN~\cite{uy2019scanobjectnn}. ModelNet40 is a synthetic dataset of CAD models containing around 10k training samples and 2.5k testing samples.
ScanObjectNN is a real-world 3D dataset with around 15k objects extracted from indoor scans. We follow the same dataset setup and preparation protocols as in ULIP~\cite{ULIP} to ensure consistent evaluation. We apply normalization on the point clouds before passing them into the input processing head and use a simple 3-layer sparse convolutional network with average pooling and 1-layer MLP output with softmax to predict the scene class. Only these two modules are trained with standard cross entropy loss on the target dataset and the 3D feature backbone is kept frozen. Results are in Tab.~\ref{tab:3dtasks} on the left.

\begin{table}[htb]
    \small
    \centering
    \vspace{-1em}
    \caption{\textbf{Left:} 3D classification results on ScanObjectNN (before slash) and ModelNet40 (after slash). \textbf{Right:} 3D segmentation results (mIOU) on ScanNet and S3DIS. For both tasks, \ConDense~outperforms all the baselines including train-from-scratch methods and pre-training methods.\ConDense~outperforms all the baselines including train-from-scratch and pre-training methods.  %Our best result outperforms SOTA largely by around 3\% on Overall Acc. 
    }
    \begin{subtable}[c]{0.53\textwidth}
    \resizebox{\columnwidth}{!}{%
    \begin{tabular}{lcc}
    \toprule
         Model& Overall Acc & Cls-mean Acc \\
         \midrule
         PointNet \cite{qi2017pointnet} &  68.2 / 89.2 & 63.4 / 86.0 \\
         %\hline
         PointNet++ \cite{qi2017pointnet++} &  77.9 / 90.7 & 75.4 / -- -- \\
         %\hline
         DGCNN \cite{wu2018dgcnn} &  78.1 / 92.9 & 73.6 / 90.2 \\
         %\hline
         MVTN \cite{hamdi2021mvtn} &  82.8 / 93.8 &  -- -- / 92.0\\
         PointMLP \cite{ma2022rethinking} &  85.7 / 94.1 & 84.4 / 91.3 \\
         %\hline
         PointNeXt \cite{qian2022pointnext} &  87.5 / -- -- & 85.9 / -- -- \\
         %\hline
         \midrule
         Point2Vec \cite{zeid2023point2vec} &  87.5 / 94.8 & 86.0 / 92.0 \\
         %\hline
         ULIP (w/ PointMLP) \cite{ULIP} &  88.8 / 94.3 & 87.8 / 92.3\\
         %\hline
         ULIP-2 (w/ PointNeXt) \cite{ULIP2} &90.8 / -- --  & 90.3 / -- -- \\
         %\hline
         ReCon \cite{qi2023contrast} &90.6 / 94.7 & -- -- / -- -- \\
         PointGPT \cite{chen2023pointgpt} &  93.4 / 94.9 & -- -- / -- -- \\
         %\hline
         \ConDense~(Ours) &  \textbf{94.1 / 95.2} & \textbf{93.4 / 93.1} \\

         %\hline
         \bottomrule
    \end{tabular}
    }
    \end{subtable}
        \begin{subtable}[c]{0.46\textwidth}
        \resizebox{\columnwidth}{!}{%
        \begin{tabular}{lcc}
    \toprule
         Model& ScanNet & S3DIS \\
         \midrule
         PointNet++ \cite{qi2017pointnet++} &  53.5 & 54.5 \\
         %\hline
         MinkowskiNet \cite{choy20194d} &  72.2 & 65.4 \\
         %\hline
         PointCNN \cite{li2018pointcnn} &  -- & 65.4 \\
         %\hline
         KPConv \cite{thomas2019kpconv} &  69.2 & 70.6 \\
         %\hline
         PointNeXt \cite{qian2022pointnext} &  71.5 & 74.9 \\
         %\hline
         PointMetaBase \cite{lin2022meta} &  72.8 & 77.0 \\
         %\hline
         PointVector \cite{deng2023pointvector} & -- & 78.4 \\
         \midrule
         PointContrast \cite{xie2020pointcontrast} &  74.1 & -- \\
         %\hline
         MSC (w/ SparseUNet) \cite{wu2023masked} &  75.5 & -- \\
         %\hline
         PPT (w/ SparseUNet) \cite{wu2023towards} &  76.4 & 78.1 \\
         %\hline
         PonderV2 (w/ SparseUNet) \cite{zhu2023ponderv2} &  77.0 & 79.9 \\
         %\hline
         Swin3D \cite{yang2023swin3d} &  77.5 & 79.8 \\
         %\hline
         \ConDense~(Ours) &  \textbf{79.8} & \textbf{80.7} \\

         %\hline
         \bottomrule
    \end{tabular}
    }
    \end{subtable}
    \label{tab:3dtasks}
\end{table}

\noindent\textbf{3D Segmentation} We follow the testing protocols in the previous works~\cite{yang2023swin3d,qian2022pointnext} to evaluate on the ScanNet~\cite{dai2017scannet} and the S3DIS~\cite{armeni2016s3dis} datasets. ScanNet contains 1613 indoor scans with 20 semantic classes, we train on its train split and report the mean Intersection over Union (mIoU) on the validation split. S3DIS contains 272 scenes, we train on its training split and evaluate on its validation set with the 6-fold cross-validation scheme. The voxel sizes for ScanNet and S3DIS are set to $2\texttt{cm}$ and $5\texttt{cm}$, respectively. We extract the point feature from the backbone with trilinear interpolation (Eq.~\ref{eq:f_3d}) and use a simple linear layer with softmax to predict the point label. Only the input processing head and the linear layer are trained with standard cross entropy loss on the target dataset and the 3D feature backbone is kept frozen. Results are presented in Tab.~\ref{tab:3dtasks} on the right.

Summarizing the results in Tab.~\ref{tab:3dtasks}, our method has demonstrated superior performance compared to other train-from-scratch and pre-training frameworks on both 3D classification and segmentation tasks. Despite only tuning the input and output heads, our approach still surpasses the performance of pre-trained methods that fine-tune the entire model. Furthermore, while many other methods heavily utilize point cloud data during pre-training~\cite{ULIP,ULIP2}, our approach achieves remarkable results without this requirement.

\subsection{2D Tasks}
To evaluate the performance of the pre-trained 2D feature backbone $\mathcal{G}_\twod$, we follow the settings as presented in DINOv2~\cite{DINOv2}, and compare with common self-supervised pre-trained baselines including MAE~\cite{he2022masked}, DINO~\cite{DINO,DINOv2}, and iBOT~\cite{zhou2021ibot}, as well as weakly supervised visual-language pre-trained model OpenCLIP~\cite{openClip}. For both classification and segmentation, we present results under the ``Linear (lin.)'' setting~\cite{DINO,DINOv2}. We include more results under the ``multi-scale (+ms)'' setting and more 3D benchmarks in our appendix. Results are presented in Tab.~\ref{tab:2dtasks}.

\noindent\textbf{2D Classification} We test the quality of the holistic image representation produced by the model on the ImageNet-1k~\cite{russakovsky2015imagenet} and Places205~\cite{zhou2014places205} classification dataset. A linear probe is added on top of the frozen feature backbone to generate the prediction, following previous works~\cite{DINO,DINOv2}.

\noindent\textbf{2D Segmentation} We test on the task of semantic image segmentation to evaluate the quality of our learned representation. A linear layer is trained to predict class logits from patch tokens and it is upscaled to obtain the final segmentation map, following previous works~\cite{DINO,DINOv2}.

\begin{table}[htb]
\vspace{-1em}
    \footnotesize
    \centering
    \caption{2D classification and segmentation results on multiple evaluation datasets with frozen features. \ConDense~improves over the DINOv2 in all benchmarks.
    }
     \resizebox{0.55\columnwidth}{!}{%
    \begin{tabular}{@{}lcccccc@{}}
    \toprule
    & \multicolumn{2}{c}{Classification (Acc)} & \multicolumn{2}{c}{Segmentation (mIOU)} \\
    \cmidrule{2-3} \cmidrule{4-5}
         Model& ImageNet & Places205 & ADE20k & PascalVOC \\
         \midrule
         OpenCLIP \cite{openClip} &  86.2 & 69.8 & 39.3 & 71.4 \\
         %\hline
         \midrule
         MAE \cite{he2022masked} &  76.6 & 52.4 & 33.3 & 67.6\\
         %\hline
         DINO \cite{DINO} &  79.2 & 60.4 & 31.8 & 66.4 \\
         %\hline
         iBOT  \cite{zhou2021ibot}  &  82.3  & 64.4  & 44.6 & 82.3 \\
         DINOv2 \cite{DINOv2} &  86.5 & 67.5 & 49.0 & 83.0 \\
         \midrule
         \ConDense &  \textbf{89.6} & \textbf{70.2} & \textbf{53.6} & \textbf{85.1} \\

         %\hline
         \bottomrule
    \end{tabular}
    }
    \label{tab:2dtasks}
\end{table}

For both 2D classification and segmentation benchmarks, our 3D-informed {\ConDense} shows consistent improvement over the original DINOv2 and indicate that our 2D-3D consensus training pipeline can help improve the performance of the existing 2D foundation models 

\subsection{Cross-Modality Scene Query}
\label{sec:exp_xmod}

Leveraging the joint 2D-3D co-embedding property of {\ConDense}, we are able to query across modalities. Here we tackle a series of matching tasks including scene identification from 2D images and a newly proposed task -- 3D scene duplication detection. The results are presented in Tab.~\ref{tab:2d3dtasks}. Dataset and test details are covered in the Appendix. We also include more experiments of 2D-3D joint query, including instance retrieval in real-world 3D scenes with 2D exemplars (\eg CAD renderings), and query with natural languages in the Appendix.

\noindent\textbf{3D Scene Retrieval with a Single Image (2D-3D).} In this task, we retrieve a scene from a repository with a single view. To compare with 2D-only methods, we first render 5 views (Ren5) from a scene and compute the cosine similarity of the query image and rendered views, and then use the winner-take-all scheme to identify the scene. In Tab.~\ref{tab:2d3dtasks}, it can be seen {\ConDense} 2D is a strong baseline not only outperforming all the other 2D methods but also performing better than ULIP-2. In 2D-3D methods, both global (using globally averaged feature) and KP (key point matching with RANSAC) variants of {\ConDense} do better than the other 2D-3D method.

\begin{table}[htb]
    \footnotesize
    \centering
    \caption{3D scene retrieval and 3D scene duplicate detection results on multiple datasets with frozen features. The upper includes 2D solutions where scenes are represented by 5 random views (Ren5), and the lower includes 2D-3D native methods.
    }
    \resizebox{0.55\columnwidth}{!}{%
    \begin{tabular}{@{}lcccccc@{}}
    \toprule
    & \multicolumn{2}{c}{3D Retrieval (Acc)} & \multicolumn{2}{c}{3D Dup. Det. (AP$_{75}^*$)} \\
    \cmidrule{2-3} \cmidrule{4-5}
         Model& Objectron & ScanNet & ScanNet & Replica \\
         \midrule
         OpenCLIP (Ren5) \cite{openClip} &  90.3 & 49.8 & 51.0 & 52.7\\
         DINOv2 (Ren5) \cite{DINOv2} &  88.1 & 43.1 & 41.3 & 43.3\\
         Unicom (Ren5) \cite{an2023unicom} &  92.9 & 52.5 & 54.3 & 57.0\\
         \ConDense~2D (Ren5) &  \textbf{94.6} & \textbf{53.3} & \textbf{58.7} & \textbf{59.0} \\
         \midrule
         ULIP-2 \cite{ULIP2} &  89.7 & 61.7 & 63.0 & 66.6\\
         \ConDense-Global &  91.6 & 70.1 & 65.3 & 66.9 \\
         \ConDense-KP & \textbf{92.9} &  \textbf{78.4} & \textbf{70.7} & \textbf{72.0} \\

         %\hline
         \bottomrule
    \end{tabular}
    }
    \label{tab:2d3dtasks}
\end{table}

\noindent\textbf{3D Scene Duplicate Detection (3D - 3D).} We further test the matching capabilities of {\ConDense} at the scene level by proposing a new task of detecting duplicate scenes in a large NeRF repository. Our method is generalizable to both NeRF and Point-Cloud inputs, and we run our experiments on ScanNet and Replica~\cite{dai2017scannet,replica19arxiv}. Results are presented in Tab.~\ref{tab:2d3dtasks}. Here, Ren5 methods are similarly defined as in the previous 3D scene retrieval, where scenes are rendered into images and the image embeddings are used. We use $\theta=0.75$ as the threshold to determine if two embeddings belong to the same scene. Here we find the remarkable effectiveness of our key points. There is a large gap between 2D-3D methods for this task, showing the need to tackle this task in 3D feature space. While \ConDense-Global performs similarly to ULIP-2, \ConDense-KP is significantly better for scene-to-scene matching.

\section{Ablation Studies and Discussions}
\label{sec:abl}

\begin{table}[htb]
    \footnotesize
    \centering
    \caption{Ablation study on removing individual components in our pre-training pipeline, evaluated on both 3D classification (Overall Acc on ScanObjectNN, before slash) and 2D classification (Acc on ImageNet-1k, after slash).}
        \resizebox{0.75\columnwidth}{!}{%
    \begin{tabular}{lcccc}
    \toprule
         Full Model\ \ \ \ \ \ \ \   &Freeze 2D\ \ \ \ \ \ \ \    &No $\mathcal{L}_\texttt{p}$\ \ \ \ \ \ \ \   &No $\mathcal{L}_\texttt{fid}$\ \ \ \ \ \ \ \  &10\% Data\ \ \ \ \ \ \ \   \\
         \midrule
         \textbf{94.1} / \textbf{89.6} &91.3 / 86.5 &90.5 / 87.1 &89.7 / 79.9 &88.7 / 79.1\\
         %\hline
         \bottomrule
    \end{tabular}
    }
    \label{tab:abl}
\end{table}

In this section, we perform experiments to verify the effectiveness of our design. The ablation study results are presented in Tab.~\ref{tab:abl}. \textbf{Freeze the 2D encoder?} When we freeze the 2D encoder, as is done in other methods~\cite{OpenScene,ULIP}, we observed a worse performance in performance for both 2D and 3D tasks. We can also see that the features from our backbone are more 3D consistent and contains more detail. Please check the visualization in our supplementary materials. 
\textbf{Sparse feature helps?} The sparse feature module is an integral component of our framework, not only enabling novel capabilities in 2D-3D retrieval but also serving as a strong self-supervision signal to enhance performance on individual 2D and 3D tasks. 
\textbf{2D fidelity helps?} The 2D fidelity loss helps prevent the 2D features from collapsing to a trivial solution or overfitting the biased data distribution. The exclusion of the 2D fidelity module has a detrimental impact on the quality of both 2D and 3D tasks, as evidenced by our experiments. Part of this loss is due to multi-view datasets being still considerably smaller than 2D image datasets, and featuring mostly man-made objects and indoor scenes. The limited size and biased distribution can cause the features to deviate significantly, leading to worse results.

\section{Conclusions}

In this work, we have presented {\ConDense}, a framework for 3D pre-training that adeptly harmonizes 2D and 3D feature extraction using pre-trained 2D networks and multi-view datasets, in both the dense and the sparse feature regimes. Our approach not only provides a pre-trained 3D network acing in multiple tasks but also enhances the quality of 2D feature representation and establishes a unified embedding space for multi-modal data interaction. Extensive experiments demonstrate the superiority of {\ConDense} over existing 3D pre-training methods in tasks like 3D classification and 3D segmentation and in new applications such as 2D image queries of 3D NeRF scenes. {\ConDense} marks an advance in 3D computer vision, reducing the reliance on scarce 3D data and enabling more efficient and multi-modality queries on 3D scenes.

Nonetheless, the pipeline comes with several limitations including the relatively high cost of processing multi-view data. Exploring more efficient ways to exploit multi-view data as well as ready-to-use 3D data could be a useful future direction. Furthermore, combining techniques from contrastive learning methods~\cite{xie2020pointcontrast,he2020momentum} and efficient fine-tuning methods (\eg LoRA~\cite{hu2021lora}) could improve the overall robustness and efficiency of the pipeline. 
We believe that {\ConDense} opens up exciting avenues for model pre-training and 2D/3D feature backbones, and defer these to future research in this direction.

\section*{Acknowledgements} 

Our thanks go to Hao-Ning Wu and Bhav Ashok for their support in building large-scale pose estimation and NeRF pipeline for dataset pre-processing.

{
    \small
    \bibliographystyle{splncs04}
    \bibliography{main}
}

\cleardoublepage
\appendix
\setcounter{page}{1}
% \renewcommand{\printtoctitle}[1]{{\ConDense} - Supplementary Material}
% \setcounter{section}{0}
% \addtocontents{toc}{\protect\setcounter{tocdepth}{3}}
% \tableofcontents

\begin{figure*}[t]
  \centering
  \includegraphics[width=0.9\textwidth]{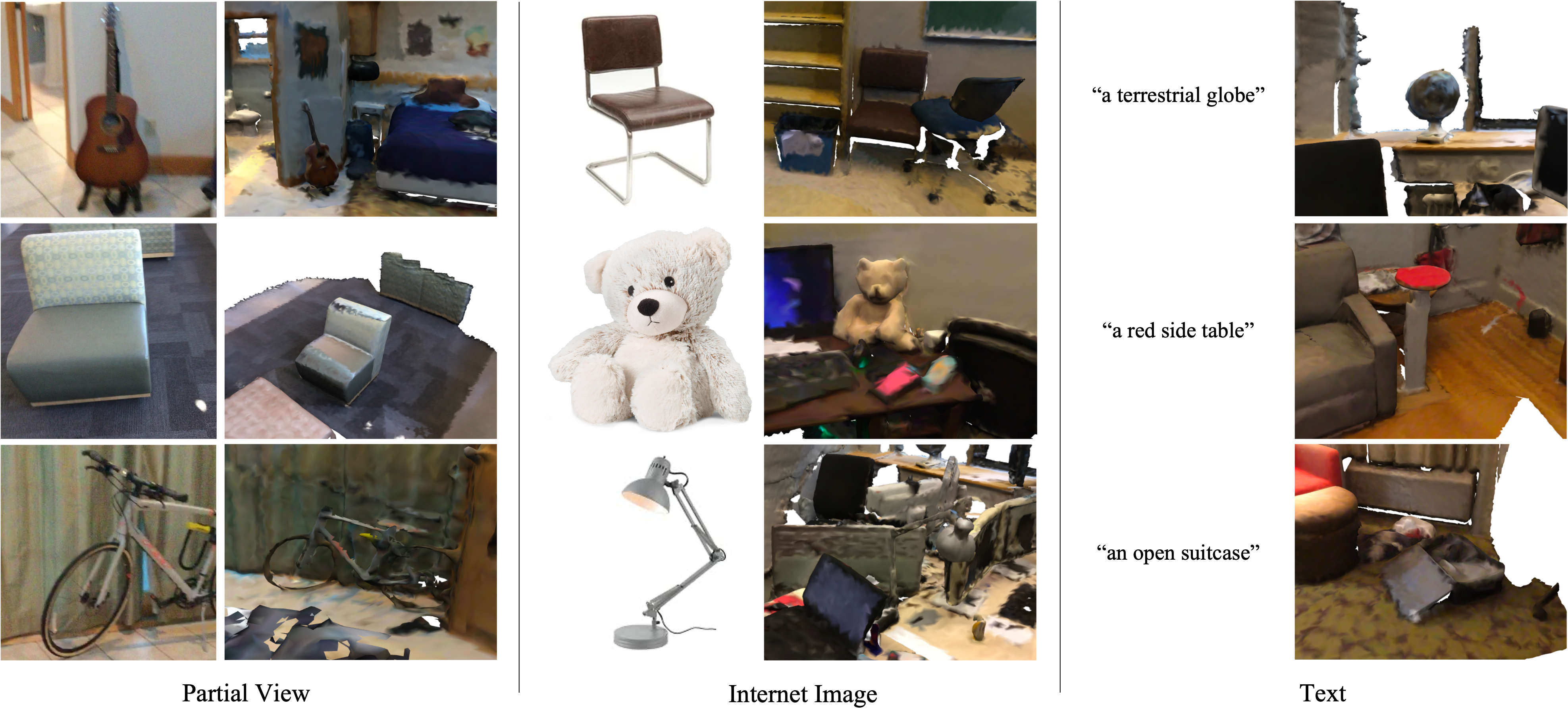}
   \caption{Visualization of using different types of input to query the target scene repository (ScanNet). Within each pair are query inputs (left) and top-1 query results (right).}
   \label{fig:supp_queries}
\end{figure*}

\section{Additional Details}

\subsection{Dataset Details}
\subsubsection{Generating NeRFs}
We use MVImgNet~\cite{yu2023mvimgnet}, ScanNet~\cite{dai2017scannet}, and RealEstate10k~\cite{zhou2018realestate10k} for our 2D-3D pre-training. To generate NeRFs for our pre-training dataset, we use the MipNeRF-360~\cite{barron2022mipnerf360} official implementation to fit the scenes. For MVImgNet, we trained 4000 steps for each scene. For ScanNet and RealEstate10k we trained 8000 steps for each scene. We half the image resolution before using it as training supervision. All the NeRF are fitted with 8 V100 GPUs. We find these settings are enough for generating NeRF with good qualities. We summarized the NeRF quality in terms of PSNR and SSIM in Tab.~\ref{tab:supp_psnr}. It takes around 35000 V100 GPU hours to build these NeRFs, and we parallelize the process on around 1000 GPUs. For ScanNet, we are determining the bounding boxes with the ground-truth meshes. For MVImgNet and RealEstate10k, we use the ray near-far values and camera locations to determine the scene bounds.

We acknowledge that fitting NeRFs on multi-view datasets requires significant computation. However, we believe this upfront cost is justified: (1) It enables 3D backbone pre-training without explicit 3D supervision, which is much more \emph{time-consuming} to acquire. (2) Creating these datasets is a one-time process, and they can be shared among researchers to avoid repeated computation. (3) Faster rendering models have emerged since we developed the pipeline, especially the 3D Gaussian Splatting, which is even more suitable for our pipeline due to its sparsity nature. Adoption of these newer models could potentially cut the computational cost greatly.

\begin{table}[htb]
    \small
    \centering
    \caption{NeRF quality on the pre-training datasets with PSNR (before the slash) and SSIM (after the slash), tested on the sampled $1\%$ images on each dataset's native image resolution.}
    \label{tab:supp_psnr}
    \begin{tabular}{lcc}
    \toprule
         MVImgNet & ScanNet & RealEstate10k \\
         \midrule
         30.23 / 0.939 & 25.78 / 0.901 & 27.24 / 0.922\\
         %\hline
         \bottomrule
    \end{tabular}
\end{table}

\subsubsection{2D Input Details} 
For MVImgNet~\cite{yu2023mvimgnet}, we use all 6.4M images in 214k scenes spreading over 238 classes for pre-training. Images are center-cropped and resized to $336\times336$ before being fed into 2D branch input. For ScanNet~\cite{dai2017scannet}, we use all 1513 scans for training. We filter out blurred frames by calculating the variance of the Laplacian matrix and ignore them for the 2D branch inputs. Other images are randomly cropped in $336\times336$ before being used as 2D inputs. The same sampling and pre-processing schemes are applied for RealEstate10k~\cite{zhou2018realestate10k}. For 2D fidelity loss, we sample from all 14M images in ImageNet-21k spreading over 21k classes. Images are resized in $336 \times 336$ before being processed by our 2D branch and output features were compared with the DINOv2 ViT-g/14 model~\cite{DINOv2} to enforce the 2D fidelity.

\subsubsection{3D Input Details}
During training, we grid sample the NeRFs with resolution varying from $64\times64\times64$ to $256\times256\times256$ to ensure adaptability to different resolutions. These samples are then re-scaled to $128\times128\times128$, with $0.5$ possibility of being sparse-dilated. The output 3D feature grid $\bfF^s$ has the spatial resolution of $128\times128\times128$ in all experiments in this paper.

\subsection{Network Details}
\subsubsection{3D Networks}
We follow the conventions and implementations as in MinkowskiNet~\cite{choy20194d} for all our 3D networks. Specifically, for input processing heads $\mathcal{J}_\texttt{3D}$, we apply 3 sparse convolution layers with ``$5\times5\times5\times1, 16$'', ``$5\times5\times5\times1, 32$'' and ``$5\times5\times5\times1, 64$'' configurations, similarly following the input processing of ~\cite{choy20194d}. Here $\times$ indicates a hypercubic sparse kernel. For the 3D reasoning backbone $\mathcal{H}_\texttt{3D}$, we apply the same architecture as MinkowskiUNet32 in ~\cite{choy20194d} but remove the original input head and modify the number of output channels to match the 2D feature channels.

\subsubsection{Key Point Prediction}
The process is illustrated in Fig.~\ref{fig:method_keypoint}. Two different 2-layer MLPs are used for reasoning key point possibilities from 2D and 3D inputs.

\subsection{Experiment Details}
\subsubsection{Computational Footprints}
For data preparation, we use V100 GPUs for fitting each scene. It takes around 35k V100 GPU hours. For pre-training, we use A100 GPUs and it takes around 4k A100 GPU hours, including both distillation and joint tuning stages. The total estimated power consumption is 12.1 MWh and carbon emitted is 5.8t CO2eq.

\subsubsection{Cross-Modality Scene Query}
We use 3 datasets: Objectron~\cite{uy2019scanobjectnn}, ScanNet~\cite{dai2017scannet}, and Replica~\cite{replica19arxiv} for cross-modality scene query tasks. To NeRF these scenes, we trained 4000 steps for Objectron and 8000 steps for both ScanNet and Replica on each scan. We use ground-truth bounding boxes included in these datasets. For ``Ren5'' baselines, we render images in the same resolution as in their corresponding datasets and the 5 views are randomly sampled from the original camera trajectories. So these 2D-Native methods are taking advantage of that queries and indices are drawn from the same trajectory, where in real-world cases this is not possible-- since the original image sequences and trajectories are typically not accessible in 3D models. For ULIP-2~\cite{ULIP2}, we use its global scene embedding to perform the top-1 matching between the queries and indices. For our methods, we use 32 key points for each 2D image and 32, 64, and 64 key points for 3D input from Objectron, ScanNet, and Replica respectively. For our 2D-3D key point matching, we use the threshold 0.75 and select the 3D scene with the most number of successful matches as the query result. 

\boldpara{3D Scene Retrieval with a Single Image (2D-3D).}
For Objectron~\cite{uy2019scanobjectnn}, we sampled 1000 scenes spreading over 9 object categories. For each scene, we sample one image as the test query. We use top-1 to match between queries and keys and calculate the retrieval accuracy. For ScanNet~\cite{dai2017scannet}, we use all 1513 scans. We sample one frame as a test query per 100 frames and at least one frame for every scene regardless of its length. 

\boldpara{3D Scene Duplicate Detection (3D - 3D).}
For both ScanNet~\cite{dai2017scannet} and Replica~\cite{replica19arxiv}, we sample 300 scene pairs where half of them are NeRFs from the same scene (duplicates) and half of them not. For ScanNet, the duplicates are created from different scans of the same scenes. (\eg scene \#0 has 3 different scans.) For Replica, the duplicates are created from the overlapping scans of the same scenes, where at least 50\% of trajectory overlappings are ensured. In this way, we comprehensively test the 3D matching capability of models on either full scans or adjacent partial scans. We use 0.75 as the threshold when determining if two embeddings belong to the same scene as a way to detect duplicated scenes. AP$_{75}^*$ is calculated as the classification accuracy of duplicate detection when the threshold is 0.75.

\subsubsection{Training and Loss Scheduling}
After initializing the 2D branches and fine-tuning the 2D key point detector, we train for 30k iterations with $\lambda_\texttt{fid}$ and $\lambda_\texttt{p} = 0$ while only tuning the 3D networks (distillation from 2D to 3D), where we sample 8 NeRF scenes in each iteration and sample rays within the original set of rays for supervision. After this stage, we train an additional 30k iterations with a linear warm-up of $\lambda_\texttt{fid}$ and $\lambda_\texttt{p}$ in 5k iterations where all 2D and 3D modules are jointly fine-tuned. Throughout the process, we use an AdamW optimizer, an initial LayerScale value of 1e-5, a weight decay cosine schedule from 0.02 to 0.24, a learning rate of 3.3e-4, and its warm-up of 2k iterations.

\section{Additional Experiments and Ablations}
\subsection{Feature Visualization}
\begin{figure*}[t]
  \centering
  \includegraphics[width=1.0\textwidth]{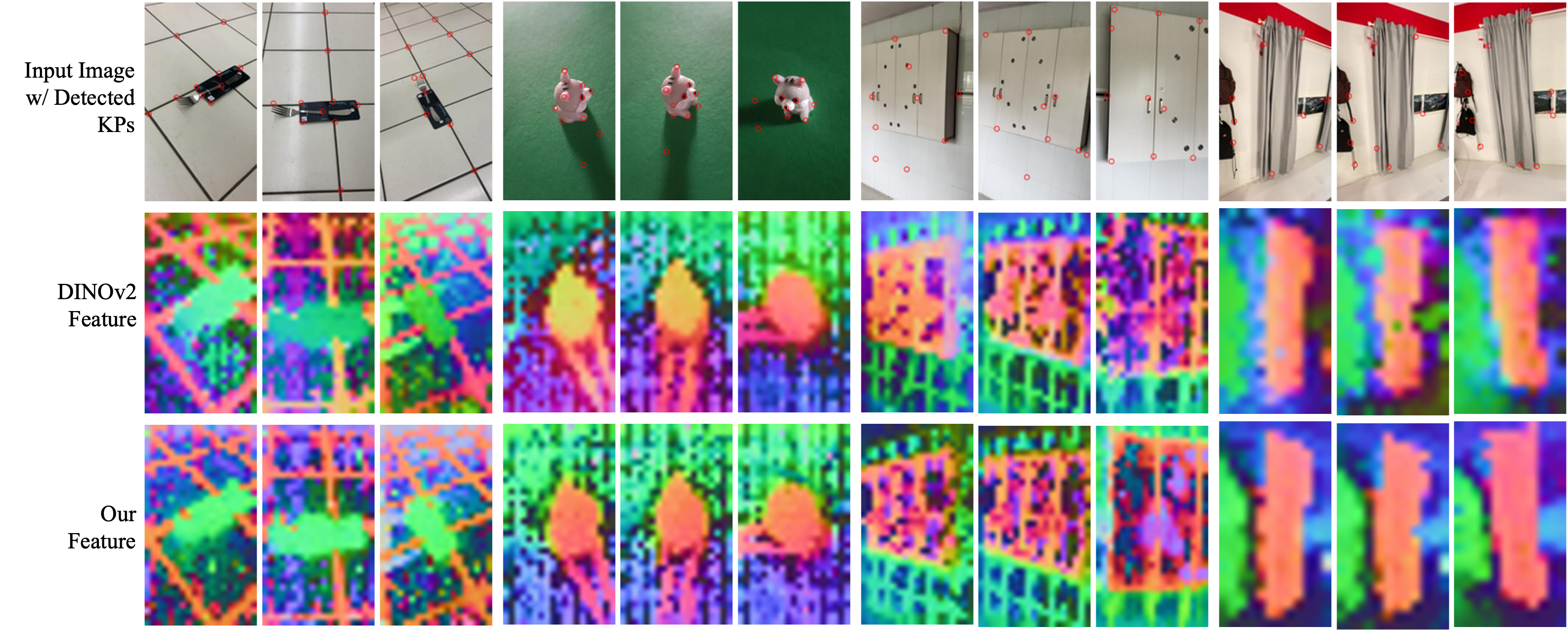}
   \caption{Visualization of our 2D dense feature reveals its superiority over the Original DINOv2 feature in terms of consistency across multi-view images. Additionally, we present visualizations of sparse feature locations identified by our key point detector. }
   \label{fig:results}
\end{figure*}
For Fig.~\ref{fig:results}, we ran PCA on the generated feature maps, both DINOv2 (2nd row) and ours (3rd row), of each scene and selected top-3 components to visualize them as red, green, and blue, respectively. We also visualize the top 10 confident (in terms of predicted probabilities) key points with positive 3D matches in this figure (1st row).

\subsection{Cross-Modality Queries}
We show in Fig.~\ref{fig:supp_queries} the visualization of using different types of input to query the target 3D scene repository (ScanNet). The unique 2D-3D co-embedded embedding space with sparse key point design enables {\ConDense} to effectively query objects in repositories of large scenes. Here, {\ConDense} is not only able to query 3D scenes with partial views from the dataset but also able to find objects in these scenes that match the appearance from unseen internet images. In addition, by changing our backbone to a multi-scale CLIP~\cite{clip} feature as in~\cite{kerr2023lerf}, we further acquire the ability to query 3D scenes with natural languages inputs, and thus build a language-image-3D co-embedded feature space with sparse key points. With this modified model, we can query scene repositories with text inputs as in Fig.~\ref{fig:supp_queries} third row.

\subsection{Additional 3D Experiments}
In the main paper, we have done experiments on 3D classification and 3D segmentation and shown our performance on multiple datasets. Here we present additional 3D experiments on different tasks and more datasets.

\subsubsection{3D Detection}
We show that our 3D features are also useful for 3D detection by following the settings of~\cite{yang2023swin3d} to attach a state-of-the-art detection head CAGoup3D~\cite{wang2022cagroup3d} and fine-tune the entire network for fair comparisions. The results are presented in Tab.~\ref{tab:3ddet}. We provide an additional 2.1 and 2.9 points gain in terms of mAP@0.25 and mAP@0.5 respectively over an already strong baseline CAGroup3D. Our performance is also better than other pre-training methods~\cite{xie2020pointcontrast,yang2023swin3d}.

\begin{table}[htb]
    \small
    \centering
    \caption{3D detection results (mAP@0.25 and mAP@0.5) on ScanNet.}
    \label{tab:3ddet}
    \begin{tabular}{lcc}
    \toprule
         Model& mAP@0.25 & mAP@0.5 \\
         \midrule
         RepSurf \cite{ran2022surface} &  71.2 & 54.8 \\
         SoftGroup \cite{vu2022softgroup} & 71.6 & 59.4 \\
         CAGroup3D \cite{wang2022cagroup3d} & 75.1 & 61.3\\
         \midrule
         PointContrast \cite{xie2020pointcontrast}& 59.2 &37.3 \\
         %\hline
         Swin3D~\cite{yang2023swin3d} w/ CAGroup3D  &  76.4 & 63.2 \\
         %\hline
         \ConDense~ w/ CAGroup3D &  \textbf{77.2} & \textbf{64.2} \\
         %\hline
         \bottomrule
    \end{tabular}
\end{table}

\begin{table}[htb]
    \small
    \centering
    \caption{2D retrieval results (top-1 Acc) on MVImgNet and ReaEstate10k datasets. }
    \label{tab:2dret}
    \begin{tabular}{lcc}
    \toprule
         Model& MVImgNet & RealEstate10k \\
         \midrule
         OpenCLIP \cite{openClip} &  70.1 & 63.0 \\
         MAE \cite{he2022masked} & 65.4 & 55.1 \\
         DINO \cite{DINO}& 65.6 & 58.9\\
         DINOv2~\cite{DINOv2}  &  70.1 & 61.3 \\
         %\hline
         \ConDense~ &  \textbf{76.9} & \textbf{63.2} \\
         %\hline
         \bottomrule
    \end{tabular}
\end{table}

\begin{table}[!htb]
\small
\centering
\caption{3D Classification (Acc) on MVImgNet, Co3D, and ShapeNet. Our method is even more useful when 3D training data are scarce (ShapeNet 1\%).
} % \caption
\label{tab:3dcls1}
% \resizebox{\linewidth}{!}{ %< auto-adjusts font size to fill line
\begin{tabular}{lccccc}
\hline
Method                                            & MVImgNet  & Co3D & ShapeNet (1\%) & ShapeNet (10\%) & ShapeNet (Full)\\ \hline
From Scratch                     &73.5 &81.1&61.2&76.9&84.9\\
\hline
PointContrast~\cite{xie2020pointcontrast}                      &76.9&84.9&65.8&78.9&86.6\\
Point-MAE~\cite{PointMAE}                     &79.1&86.2&72.3&81.1&89.2\\
ULIP-2~\cite{ULIP2}                     &82.9&86.1&74.3&81.9&89.3\\
Ours (Freeze 2D)                     &87.3&88.8&80.1&83.9&90.1\\
Ours                           & \textbf{91.3}& \textbf{93.8}& \textbf{81.4}& \textbf{85.3}& \textbf{91.9}\\  \hline
\end{tabular}
% } %< \resizebox
\end{table}

\begin{table}[htb!]
\centering
\small
% \resizebox{\linewidth}{!}{ %< auto-adjusts font size to fill line
\caption{3D Segmentation (mIOU) on ScanNet, SemanticKitti, and S3DIS. Our method is even more useful when 3D training data are scarce (ScanNet 1\%).
} % \caption
\label{tab:3dseg1}
\begin{tabular}{lccccc}
\hline
Method                                            & ScanNet (1\%) & ScanNet (10\%) & ScanNet (100\%) & SemanticKitti  & S3DIS \\ \hline
PointNet~\cite{qi2017pointnet}                     &42.2 &62.1&72.2&19.6&47.6\\
Mix3D~\cite{nekrasov2021mix3d}                       &39.4 &69.9&73.6&65.4&63.5\\
\hline
PointContrast~\cite{xie2020pointcontrast}                      &52.9&70.4&74.1&71.7&75.2\\
Swin3D~\cite{yang2023swin3d}                     &54.8&65.2&77.5&74.7&79.8\\
Ours (Freeze 2D)                     &65.6&70.0&78.1&74.6&80.6\\
Ours                           & \textbf{67.3}& \textbf{72.3}& \textbf{79.8}& \textbf{75.1}& \textbf{80.7}\\  \hline
\end{tabular}
% } %< \resizebox
\end{table}

\subsubsection{More on 3D Classification and 3D Segmentation}
We tested our 3D capabilities on more datasets and also with varying amounts of 3D training data. The results are presented in Tab.~\ref{tab:3dcls1} (classification) and Tab.~\ref{tab:3dseg1} (segmentation). We see consistent improvement over all baselines from these results. Also, it can be seen that our method is even more useful when 3D training data is very scarce (See ShapeNet 1\% and ScanNet 1\%).

\subsection{Additional 2D Experiments}
In the main paper, we have done experiments on 2D classification and 2D segmentation and shown our performance on multiple datasets. Here we present additional 2D experiments on more tasks.

\subsubsection{2D Retrieval}
We evaluate our performance for image retrieval with MVImgNet~\cite{yu2023mvimgnet} and ScanNet~\cite{dai2017scannet}. We follow the experiment settings of~\cite{DINO,DINOv2} by freezing the features and
directly applying k-NN for retrieval. On both datasets, we perform tests with 1 query image and 1 index image on each scene. The top-1 accuracy is reported in Tab.~\ref{tab:2dret}. Our {\ConDense} clearly generates better global features suitable for retrieval tasks.

\subsubsection{Depth Estimation}
We follow the settings in~\cite{DINOv2} and attach a linear classifier on top of one (lin. 1) or four (lin. 4) transformer layers to infer depth from the frozen feature backbones. It can be seen that our performance is better than all other pre-training backbones except DINOv2. The reason is that the 2D-3D consensus loss enforces the feature to be invariant across different views, and the 2D branch is thus expected to generate the same features for the same spatial point regardless of viewing angles and distances. The process of 2D-3D consensus facilitates more 3D-informed and consistent features, as can be observed from Fig.~\ref{fig:results}. Such a property could be desired or unwanted depending on the exact downstream tasks. And we defer the more in-depth study into this property to future research. To validate this, we show our results on multi-view stereo (MVS) depth estimation. Here we tested two widely used MVS depth estimation methods-- MVSNet~\cite{yao2018mvsnet} and PointMVS~\cite{chen2019point}, on the standard dataset (DTU~\cite{jensen2014large}). It can be seen that both backbones are boosted by our features, and {\ConDense} outperforms DINOv2 by a large margin.

\begin{table}[htb]
 \footnotesize
  \centering
    \caption{Depth estimation with frozen features. We report performance when training a linear classifier on top of one (lin. 1) or four (lin. 4) transformer layers. We report the RMSE metric on the 3 datasets. Lower is better.}
  \label{tab:2dmdepth}
  \begin{tabular}{@{}ll c cc c cc@{}}
    \toprule
    & & \multicolumn{2}{c}{NYUd} && \multicolumn{2}{c}{NYUd $\rightarrow$ SUN RGBD} \\
    \cmidrule{3-4} \cmidrule{6-7}
    Method  && lin. 1 & lin. 4 && lin. 1 & lin. 4 \\
    \midrule
    OpenCLIP~\cite{openClip}  && 0.541 & 0.510   && 0.537 & 0.476 \\
    \midrule
    MAE~\cite{he2022masked}  && 0.517 & 0.483 && 0.545 & 0.523 \\
    DINO~\cite{DINO}         && 0.555 & 0.539 && 0.553 & 0.541 \\
    iBOT~\cite{zhou2021ibot}  && 0.417 & 0.387  && 0.447 & 0.435 \\
    DINOv2~\cite{DINOv2}      && \bf 0.344 & \bf 0.298 && 0.402 & \bf 0.362 \\ 
    Ours                    &&  0.361 & 0.322 &&  \bf0.389 & 0.367 \\
    \bottomrule
  \end{tabular}
\end{table}
ßß
\begin{table}[htb]
    \small
    \centering
    \caption{Stereo depth estimation on the DTU dataset. Backbone methods could benefit a lot from our feature initialization.}
    \label{tab:2dsdepth}
    \begin{tabular}{lc}
    \toprule
         Model& Overall Err. \\
         \midrule
         MVSNet \cite{yao2018mvsnet} &  0.462  \\
         PointMVS \cite{chen2019point} & 0.366  \\
         \midrule
         DINOv2~\cite{DINOv2} w/ MVSNet &  0.389  \\
         DINOv2~\cite{DINOv2} w/ PointMVS &  0.365  \\
         \ConDense~ w/ MVSNet &  0.341  \\
         \ConDense~ w/ PointMVS &  \textbf{0.320}  \\
         %\hline
         \bottomrule
    \end{tabular}
\end{table}

\subsubsection{More on 2D Segmentation}
We test our performance on 2D segmentation following the ``+ms'' setting as proposed in~\cite{DINOv2}. The results are presented in Tab.~\ref{tab:2dsemseg1}.

\begin{table}[htb]
 \small
  \centering
    \caption{Semantic segmentation on ADE20K, CityScapes, and Pascal VOC with frozen features and a linear classifier (lin.) and with multiscale (+ms).
  }
  \label{tab:2dsemseg1}
  \begin{tabular}{@{}ll c cc c cc c cc@{}}
    \toprule
    & & \multicolumn{2}{c}{ADE20k} && \multicolumn{2}{c}{CityScapes} && \multicolumn{2}{c}{Pascal VOC} \\
    \cmidrule{3-4} \cmidrule{6-7} \cmidrule{9-10}
    Method  && lin. & +ms && lin. & +ms && lin. & +ms \\
    \midrule
    OpenCLIP~\cite{openClip}  && 39.3 & 46.0   && 60.3 & 70.3 && 71.4 & 79.2 \\
    \midrule
    MAE~\cite{he2022masked}         && 33.3 & 30.7 && 58.4 & 61.0 && 67.6 & 63.3 \\
    DINO~\cite{DINO}         && 31.8 & 35.2 && 56.9 & 66.2 && 66.4 & 75.6 \\
    iBOT~\cite{zhou2021ibot}        && 44.6 & 47.5 && 64.8   & 74.5  && 82.3 & 84.3 \\
    DINOv2~\cite{DINOv2}      && 49.0 & 53.0 && 71.3 & 81.0 && 83.0 & 86.2 \\ 
    \midrule
    Ours      && \bf 50.2 & \bf 53.8 && \bf 74.1 & \bf 83.1 && \bf 83.2 & \bf 86.5 \\
    \bottomrule
  \end{tabular}
\end{table}
ß
\subsection{Additional Ablation Studies}
\subsubsection{Effect of NeRF Quality} We observed differences in performance when using trained NeRFs of different quality. The numbers are reported on ImageNet (2D Cls) and ScanObjectNN (3D Cls) on a lighter version of the final model reported in the main paper.

\begin{table}[htb]
 \small
  \centering
      \caption{Effect of NeRF Quality}
  \label{tab:suppabl1}
\scalebox{1.0}{
\begin{tabular}{l|cccccc|c}
\hline
Training NeRF Type             & Mip-NeRF (2k steps)	&Mip-NeRF (4k steps)\\ \hline
Quality (PSNR/SSIM)    & 28.54 / 0.899     & 30.23 / 0.939 \\ \hline
2D Cls / 3D Cls       & 86.2 / 88.9   & \textbf{87.7} / \textbf{93.2}   \\
\hline
\end{tabular}
}
\end{table}

Results are presented in Tab.~\ref{tab:suppabl1}. We find that pre-training with data of lower quality will have an impact on performance, especially for 3D tasks. We take measures such as using different iteration numbers to ensure convergence, and filtering out low-quality frames as mentioned previously.

\subsection{Commonly Used Backbones}
The MinkowskiNet (MNet) is the established state-of-the-art for dense 3D tasks and has been adopted by most 3D dense task models. So we select MinkowskiNet as our backbone. Here we had an experiment comparing different backbones -- results are reported below on ScanObjectNN (3D Cls) and ScanNet (3D Seg).

\begin{table}[htb]
 \small
  \centering
      \caption{Comparing different backbones.}
\noindent\scalebox{1}{
\begin{tabular}{l|cccccc|c}
\hline
                      & PointBERT	&PointNeXt   &MinkowskiNet \\ \hline
\#Parameters          & 32.3M    & 41.6M   & 41.3M\\ \hline
3D Cls / 3D Seg       & 87.2 / -    & 92.1 / 77.0  & \textbf{93.2} / \textbf{79.1} \\
\hline
\end{tabular}
}
\end{table}

We see better performance, especially on 3D Seg, with MNet. The reasons behind this may include \underline{\textbf{(1)}} an overfitting tendency of the transformer-based model PointBERT, which aligns with OpenScene's finding (their Sec. 6.4); \underline{\textbf{(2)}} our better generalizability from training to test domains, where point distributions are different. We will include a more detailed discussion on the performance and scalability of different backbones once time permits.

\end{document}